\documentclass{article}
\usepackage{spconf,amsmath,graphicx,bigstrut,multirow,subcaption,makecell,url}
\usepackage[colorlinks,linkcolor=red]{hyperref}

\title{Learn an Effective Lip Reading Model without Pains}
\name{Dalu Feng$^{1,2}$, Shuang Yang$^{1,2}$, Shiguang Shan$^{1,2}$, Xilin Chen$^{1,2}$}
\address{$^{1}$Key Laboratory of Intelligent Information Processing of Chinese Academy of Sciences (CAS), \\Institute of Computing Technology, Beijing, 100190, China\\
$^{2}$University of Chinese Academy of Sciences, Beijing, 100049, China}
\begin{document}
%

\maketitle
\begin{abstract}
Lip reading, also known as visual speech recognition, aims to recognize the speech content from videos by analyzing the lip dynamics. There have been several appealing progress in recent years, benefiting much from the rapidly developed deep learning techniques and the recent large-scale lip-reading datasets. Most existing methods obtained high performance by constructing a complex neural network, together with several customized training strategies which were always given in a very brief description or even shown only in the source code. We find that making proper use of these strategies could always bring exciting improvements without changing much of the model. Considering the non-negligible effects of these strategies and the 
existing tough status to train an effective lip reading model, we perform a comprehensive quantitative study and comparative analysis, for the first time, to show the effects of several different choices for lip reading. By only introducing some easy-to-get refinements to the baseline pipeline, we obtain an obvious improvement of the performance from 83.7\% to 88.4\% and from 38.2\% to 55.7\% on two largest public available lip reading datasets, LRW and LRW-1000, respectively. They are comparable and even surpass the existing state-of-the-art results. \footnote{\url{https://github.com/Fengdalu/learn-an-effective-lip-reading-model-without-pains}}
\end{abstract}
\section{Introduction}
\label{sec:intro}

Automatic lip reading aims to recognize the speech content by watching videos. It has lots of potential applications in both noisy and silent environments. However, several factors, including the lighting conditions, speaker's age, make-up, viewpoints, and so on, make lip reading a challenging task.

Fortunately, the recent progress in the following two points makes automatic lipreading possible. Firstly, the rapidly developed deep learning techniques have been proved to be able to tackle several problems which are closely related to lip reading, including action recognition\cite{wang2018non,feichtenhofer2019slowfast,qiu2017learning}, sequential modeling \cite{sutskever2014sequence,devlin2018bert,vaswani2017attention}, an so on. Secondly, several large-scale lip reading datasets have been released in recent years \cite{chung2017lip,yang2019lrw,chung2017liplrw}, which provide a huge amount of data with large variations and contribute much to the progress of lip reading. By taking full advantages of these two aspects, several appealing results have been presented recently lip reading \cite{assael2016lipnet, stafylakiscombining, martinez2020lipreading}.

Most modern deep lip reading models consist two modules: the frontend module and the backend module. 
The frontend module often pays more attention to the local motion patterns, including the frame-level and clip-level features. While the backend module focuses more on the whole sequence-level patterns and often is designed to learn the temporal dynamics of the sequence based on the output features of the frontend module. 
Although the architectures of most models can be divided into these two parts, there has never been consensus on which strategies or pipeline could bring effective learning of the lip reading model. 
Different work always have their own different strategies to obtain effective lip reading. 
For example, Stafylakis et al. \cite{stafylakiscombining} introduced a multi-stage procedure by training the frontend and the backend separately at first and then tuning them together to obtain effective lip reading. Martinez et al. \cite{martinez2020lipreading} proposed an end-to-end procedure together with a cosine learning rate scheduling to perform training. Ma et al.\cite{ma2020towards} applied the AdamW optimizer, not the traditional Adam optimizer to perform optimization. 

In this paper, we perform a comprehensive quantitative study and comparative analysis to the effects of several factors for lip reading, including the learning rate scheduling, the data pre-processing, the choices for different modules in the pipeline, and so on. 
Besides referring to the existing lip-r
eading methods, we also borrow several successful tricks from the general computer vision domain and build a new basic training pipeline for effective lip reading. 
%
Finally, without changing much of the model, we obtain comparable results and even better results than the current state-of-the-art on the largest public available word-level datasets.
\vspace{-1em}
\section{Related work}
\label{sec:related}
\vspace{-0.5em}
\subsection{Deep Lip Reading}
\vspace{-0.5em}
Research on lip-reading has a long history. Most earlier methods are based on hand-crafted features with shallow models, such as Hidden Markov Models (HMM) \cite{puviarasan2011lip}, Discrete Cosine Transform (DCT) \cite{hong2006pca}, Active Appearance Model (AAM) \cite{matthews1998lipreading}, and so on.
With the rapid developments of the deep learning techniques, researchers in this area begin to introduce deep neural networks in recent years. 
For example, Chung et al. \cite{chung2017liplrw} proposed a multi-tower CNN architecture based on VGG-M and reported the results of deep learning methods on large scale lip reading datasets for the first time. %
Stafylakis et al.\cite{stafylakiscombining, stafylakis2018pushing} introduced the deep Residual Network \cite{he2016deep} as the frontend for the first time and obtained appealing results. 
Recently, Martinez and Ma et al. \cite{martinez2020lipreading, ma2020towards} proposed to introduce a temporal convolutional neural network for the lip reading problem. 
With these impressive methods, state-of-the-art performance has been raised from 61.1\% to 87.0\% on the largest English word-level lipreading dataset LRW in only four years. However, almost all of these methods have their own customized setting in either the training procedure or data processing. So a comprehensive study and comparative analysis about which factors could really bring improvements to the lip reading task are necessary at this time.
\begin{figure}[t]
\label{Figure1}
\setlength{\abovecaptionskip}{-0.2cm}
\begin{minipage}[t]{1.0\linewidth}
  \centering
  \centerline{\includegraphics[width=8cm]{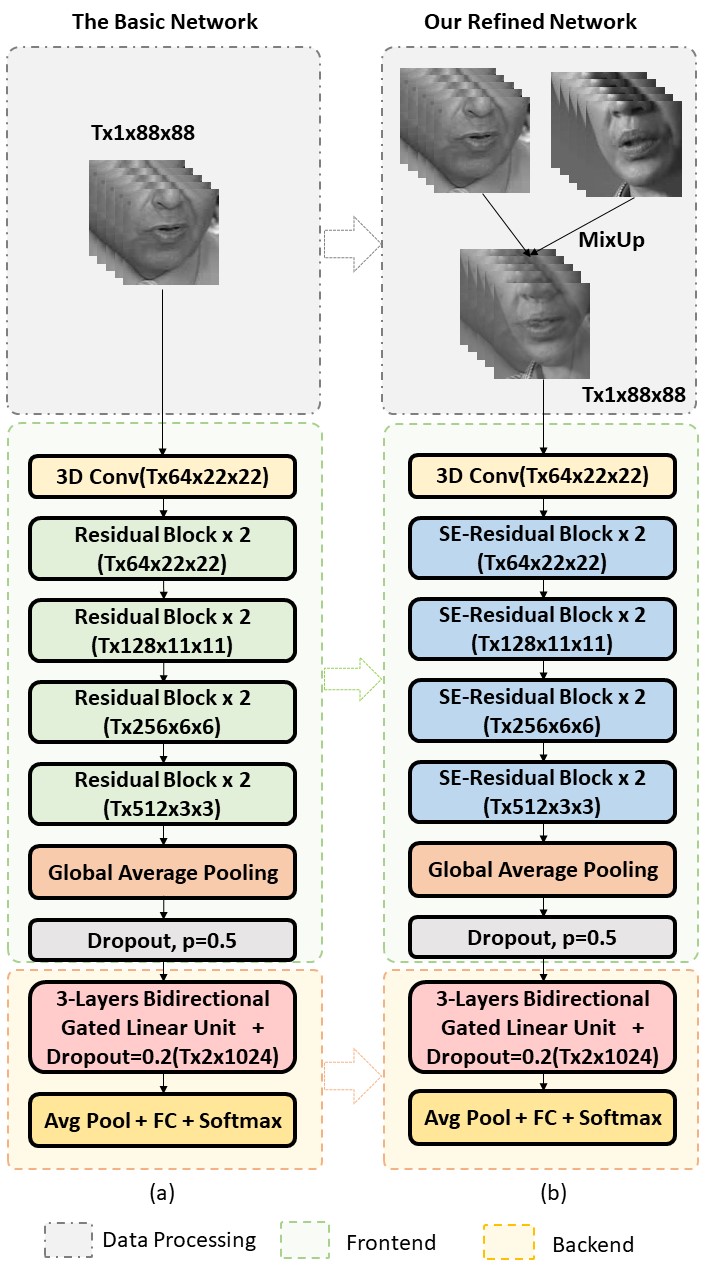}}
\end{minipage}
\vspace{-0.2cm}
\caption{Our basic pipeline (a) and our refined pipeline (b)}
\label{fig:architecture}
\end{figure}
\vspace{-0.5em}
\subsection{Bag Of Tricks in Deep Learning}
\vspace{-0.5em}
Several training heuristics and architecture refinements have been proved to be able to improve numerous kinds of tasks in the domain of computer vision. 
He et al. \cite{he2019bag} have gathered bag of tricks for training, leading to improvement of the accuracy of image classification on ImageNet. Luo et al. \cite{luo2019bag} have explored a strong and simple baseline for Person Re-identification by introducing several training tweaks. In the object detection area, there are also plenty of work focusing on the setting of training pipelines and architectures \cite{zhang2019bag,redmon2018yolov3,bochkovskiy2020yolov4}. All of these work present a common and easier way to perform the given task and provide great impetus to the follow-up developments on the target task.
\vspace{-1.2em}
\section{The Basic Pipeline}
\vspace{-0.5em}
\label{sec:pipeline}

We adopt a popular pipeline in the lip reading domain as our baseline, as shown in Fig \ref{fig:architecture}(a). ResNet-18 is used as the frontend module. The first 2D convolutional layer is modified to a 3D convolutional layer with kernel size of 5$\times$7$\times$7. Then a global average pooling is performed on the output of residual blocks and the output features are feeded to the backend network. The final fully connection layer's output dimension is equal to the total number of word classes. We adopt the following setting for training by default:
\vspace{0.5em}
\\
\indent\textbf{(1) Initialization:} The parameters of the involved convolutional modules and the recurrent modules are randomly initialized with the same manner as \cite{he2019bag}. It can be divided into three folds: (1) In convolutional layers, we set the parameters to random values uniformly drawn from $[-a, a]$, where $a = \sqrt{2/(d_{in}+d_{out})} $, $d_{in}$ and $d_{out}$ are the input and the output size. (2) In batch normalization layers, we set all $\gamma$ vectors set to 1 and all $\beta$ vectors to 0 in the affinition opration: $ y = \gamma x + \beta $, where $x$ is the normalized feature in BatchNorm and $y$ is the output of the BN layer. (3) In the Gated Recurrent Unit, all parameters are drawn from $(-1, 1)$ and parameters of the final fully connneted layer are sampled from uniform distribution $[-1, 1]$. 
\vspace{0.5em}
\\
\indent\textbf{(2) Data Processing:} We shuffle the order of the input videos at each epoch, resize them to 96$\times$96, and then random crop them to 88$\times$88 as the final input to the model. We select a batch of videos in each training iteration. Each video is flipped horizontally with probability 0.5, converted to grayscale, and normalized to [0, 1]. A special setting on LRW-1000 is that 
we chose 40 frames for each word and put the target word at the center to make it similar to the data in LRW. We found that it can provide more context , so as to improve the performance of lip reading. This setting is adopted by default in our experiments on LRW-1000.
\vspace{0.5em}
\\
\indent\textbf{(3) Loss:} In the backend, we average the GRU's outputs in the temporal dimension and send the results to the final fully connected layer for prediction. The cross-entropy loss is used for the optimization.
\vspace{0.5em}
\\
\indent\textbf{(4) Optimizer:} By default, the Adam Optimizer \cite{kingma2014adam} is adopted. The initial learning rate is 3e-4 and the weight decay is 1e-4. When the model is trained on a single GPU, the batch size is set to 32. When training on different devices, we linearly adjust the learning rate according to the batch size. 
We validate the model at the end of each epoch and the learning rate will be reduced by a factor of 2 whenever the validation error plateaus in 3 continuous epochs. The minimal learning rate is set to 1e-6.

All the experiments are performed on the LRW and LRW-1000 dataset, which are the only two public largest word-level lip reading datasets. The LRW dataset is an English lip reading dataset, with samples of 500-word classes from the BBC programmes. The LRW-1000 dataset is a Mandarin lip-reading dataset, consisting of 1000 word/phrase classes spoken by more than 2000 speakers.
\begin{table}[t]
\normalsize
\centering
    \begin{tabular}{!{\vrule width1.2pt}c|c!{\vrule width1.2pt}c|c!{\vrule width1.2pt}}
    \Xhline{1.2pt}
    \textbf{Frontend} & \textbf{Backend} & \textbf{LRW} & \textbf{LRW-1000} \bigstrut\\
    \Xhline{1.2pt}
    VGGM$^*$  & -     & 61.1\%  & 25.7\% \bigstrut\\
    \hline
    ResNet-18$^*$ & \multirow{4}[6]{*}{3 Layers GRU} & 83.0\% & 38.2\% \bigstrut\\
\cline{1-1}\cline{3-4}    ResNet-34$^*$ &       & 83.5\%  & - \bigstrut\\
\cline{1-1}\cline{3-4}    ResNet-18 &       & 83.7\%  & 46.5\% \bigstrut\\
\cline{1-1}\cline{3-4}    SE-ResNet-18 &       & 84.1\%  & 46.8\% \bigstrut\\
    \Xhline{1.2pt}
    \end{tabular}%
\caption{Comparison of different frontend modules, where $^*$ denotes that the results are from existing work.}
\label{table:frontend}
\vspace{-0.1cm}
\end{table}
\begin{table}[t]
\normalsize
\centering
    \begin{tabular}{!{\vrule width1.2pt}c|c!{\vrule width1.2pt}c|c!{\vrule width1.2pt}}
    \Xhline{1.2pt}
    \textbf{Frontend} & \textbf{Backend} & \textbf{LRW} & \textbf{LRW-1000} \bigstrut\\
    \Xhline{1.2pt}
    \multirow{4}[8]{*}{ResNet-18} & 3 Layers GRU & 83.7\%  & 46.5\%  \bigstrut\\
\cline{2-4}          & GRU w/o droppout & 83.1\%  &  45.5\%\bigstrut\\
\cline{2-4}          & MS-TCN & 83.4\%  & 43.0\%  \bigstrut\\
\cline{2-4}          & Transfomer$^*$ & 76.2\%  & 44.5\%  \bigstrut\\
    \Xhline{1.2pt}
    \end{tabular}%
\caption{Comparison of different backend modules, where $^*$ denotes that the results are from existing work.}
\label{table:backend}
\end{table}
\begin{table}[t]
  \centering
    \begin{tabular}{!{\vrule width1.2pt}c!{\vrule width1.2pt}c|c!{\vrule width1.2pt}}
    \Xhline{1.2pt}
    \textbf{Data Processing} & \textbf{LRW} & \textbf{LRW-1000} \bigstrut\\
    \Xhline{1.2pt}
    Baseline & 83.7\%  & 46.5\% \bigstrut\\
    \hline
    Aligned Lip & 84.2\% & - \bigstrut\\
    \hline
    Word Boundary Input & 86.5\% & 53.6\% \bigstrut\\
    \Xhline{1.2pt}
    \end{tabular}%
  \caption{Comparison of different data processing strategies, where the original data provided in LRW-1000
dataset has been already aligned.}
  \label{table:dataprocess}%
\end{table}%
\begin{table}[t]
  \centering
    \begin{tabular}{!{\vrule width1.2pt}c!{\vrule width1.2pt}c|c!{\vrule width1.2pt}}
    \Xhline{1.2pt}
    \textbf{Training Tweaks} & \textbf{LRW} & \textbf{LRW-1000} \bigstrut\\
    \Xhline{1.2pt}
    Baseline  & 83.7\%  & 46.5\% \bigstrut\\
    \hline
    MixUp & 84.0\%  & 47.3\% \bigstrut\\
    \hline
    Label Smooth & 84.2\%  & 47.0\% \bigstrut\\
    \hline
    Cosine Scheduler & 84.2\%  & 46.6\% \bigstrut\\
    \hline
    Exp Scheduler & 83.2\%  & 45.6\% \bigstrut\\
    \Xhline{1.2pt}
    \end{tabular}%
  \caption{Effects of different training tweaks for lip reading.}
  \label{table:training}%
\end{table}%
\vspace{-0.5em}
\section{Bag Of Tricks For Deep Lip Reading}
\vspace{-0.5em}
\label{sec:bagoftricks}
This section will introduce several tricks from different aspects, including the model refinements, training tricks,  and data processing settings. We perform a case-by-case study for each refinement in the following part.\footnote{Unless specifically noted, all the results in Tab.1-Tab.4 are performed using the the basic pipeline in Sec.3.}
\vspace{-0.5em}
\subsection{Model Refinements}
We perform comparison and analysis on the frontend and backend modules separately, to show their effects for lip reading respectively. 
\vspace{0.5em}
\\
\indent\textbf{The Frontend Network} We compare different types of frontend module with the baseline's GRU backend. Although deeper convolutional neural networks always perform better than shallow networks in general tasks, the ResNet-34 performs slightly better than ResNet-18 on LRW, as shown by the 2nd-3rd rows in Tab.\ref{table:frontend} .
But when introducing our particular setting on LRW-1000 and the learning rate schedule as described in Sec.\ref{sec:pipeline}, ResNet-18 has shown an obvious improvement on both datasets. Inspired by the success of the Squeeze-and-Extract module \cite{hu2018squeeze} in other computer vision tasks, we also introduce it to the baseline model to evaluate its effect for lip reading. As shown in the Tab. \ref{table:frontend}, the introduction of this module could lead to an stable improvement of the performance on both LRW and LRW-1000.
\vspace{0.5em}
\\
\indent\textbf{The Backend Network} We compare three popular types of backend modules in the lip-reading area: the GRU based RNN network, the Temporal Convolutional Network \cite{martinez2020lipreading}, and the Transformer. The basic pipeline in Fig.\ref{fig:architecture}(a) is used with different backend modules in this part. As shown in Tab. \ref{table:backend}, the basic pipeline with GRU based backend always performs better on both the two large datasets, with exactly the same conditions (including the same data, the same learning rate schedule, using no pre-training and so on).
\begin{table*}[!t]
  \centering
    \begin{tabular}{!{\vrule width1.2pt}c|c|c|c|c|c|c|c!{\vrule width1.2pt}}
    \Xhline{1.2pt}
    \textbf{Year} & \textbf{Method} & \textbf{Frontend} & \textbf{Backend} & \textbf{Data Type} & \textbf{Input Size} & \textbf{LRW} & \multicolumn{1}{c!{\vrule width1.2pt}}{\textbf{LRW-1000}} \bigstrut\\
    \Xhline{1.2pt}
    2016  & Chung et al.\cite{chung2017liplrw} & VGGM  & -     & Lip   & 112$\times$112 & 61.1\%  & \multicolumn{1}{c!{\vrule width1.2pt}}{25.7\%} \bigstrut\\
    \hline
    2017  & Stafylakis et al.\cite{stafylakiscombining} & ResNet-34 & \multirow{2}[4]{*}{BiLSTM} & Lip   & \multirow{2}[4]{*}{112x112} & 83.5\%  & \multicolumn{1}{c!{\vrule width1.2pt}}{38.2\%} \bigstrut\\
\cline{1-3}\cline{5-5}\cline{7-8}    2018  & Stafylakis et al.\cite{stafylakis2018pushing} & ResNet-18 &       & + Word Boundary &       & 88.8\%  & \multicolumn{1}{c!{\vrule width1.2pt}}{-} \bigstrut\\
    \hline
    2019  & \makecell[c]{Wang et al. \cite{wang2019multi}\\Multi Graned} & \makecell[c]{Multi-Grained\\ResNet-18} & \makecell[c]{Conv\\BiLSTM} & Lip   & 88$\times$88 & 83.3\%  & \multicolumn{1}{c!{\vrule width1.2pt}}{36.9\%} \bigstrut\\
    \hline
    2019  & \makecell[c]{Weng et al. \cite{weng2019learning}\\Two Stream} & \makecell[c]{Two-Stream\\ResNet-18} & BiLSTM & Lip   & 112$\times$112 & 84.1\%  & \multicolumn{1}{c!{\vrule width1.2pt}}{-} \bigstrut\\
    \hline
    2020  & \makecell[c]{Luo et al. \cite{luo2020pseudo}\\Policy Gradient} & ResNet-18 & BiGRU & Lip   & 88$\times$88 & 83.5\%  & \multicolumn{1}{c!{\vrule width1.2pt}}{38.7\%} \bigstrut\\
    \hline
    2020  &\makecell[c]{Zhao et al. \cite{zhao2020mutual}\\Mutual Information} & ResNet-18 & BiGRU & Lip   & 88$\times$88 & 84.4\%  & \multicolumn{1}{c!{\vrule width1.2pt}}{38.7\%} \bigstrut\\
    \hline
    2020  & \makecell[c]{Xiao et al. \cite{xiao2020deformation}\\ Deformation Flow} & ResNet-18 & BiGRU & Lip   & 88$\times$88 & 84.1\%  & \multicolumn{1}{c!{\vrule width1.2pt}}{41.9\%} \bigstrut\\
    \hline
    2020  & \makecell[c]{Zhang et al.\\Face Cutout} & ResNet-18 & BiGRU & Aligned Face & 112$\times$112 & 85.0\% & \multicolumn{1}{c!{\vrule width1.2pt}}{45.2\%} \bigstrut\\
    \hline
    2020  & \makecell[c]{Martinez et al. \cite{martinez2020lipreading}\\Temporal Convolution} & ResNet-18 & MS-TCN & Aligned Lip & 88$\times$88 & 85.3\%  & \multicolumn{1}{c!{\vrule width1.2pt}}{41.4\%} \bigstrut\\
    \hline
    2020  &\makecell[c]{Ma et al. \cite{ma2020towards}\\Multi-Stage Distillation} & ResNet-18 & MS-TCN & Aligned Lip & 88$\times$88 & 87.7\%  & \multicolumn{1}{c!{\vrule width1.2pt}}{43.2\%} \bigstrut\\
    \Xhline{1.2pt}
    \multicolumn{2}{!{\vrule width1.2pt}c|}{\multirow{2}[4]{*}{\makecell[c]{Ours\\SE+MixUp+Cosine LR+LS+WB}}} & \multirow{2}[4]{*}{SE-ResNet-18} & \multirow{2}[4]{*}{BiGRU} & Aligned Lip & \multirow{2}[4]{*}{88$\times$88} & 85.0\% & \multicolumn{1}{c!{\vrule width1.2pt}}{48.0\%} \bigstrut\\
\cline{5-5}\cline{7-8}    \multicolumn{2}{!{\vrule width1.2pt}c|}{} &  &  & + Word Boundary &       & 88.4\%  & 55.7\%\bigstrut\\
    \Xhline{1.2pt}
    \end{tabular}%
  \caption{Comaprison with existing methods. '+LS' and '+WB' means that label smoothing and word boundaries are included.}
  \label{tab:final}%
\vspace{-0.5cm}  
\end{table*}%
\vspace{-0.5em}
\subsection{Data Processing}
In most existing methods, the lip region is cropped by a fixed-size rectangle, and the word boundary information is often discarded. Here we perform an analysis to the effects of these two factors, as shown in Tab. \ref{table:dataprocess}.
\vspace{0.5em}
\\
\indent\textbf{Face Alignment} Face alignment is always helpful for face recognition. Recently, Zhang et al. \cite{zhang2020can} found that face alignment can also improve the accuracy of visual speech recognition. Inspired by their work, we perform face alignment at first before the lip region extraction. We use dlib toolkit \cite{king2009dlib} to get facial landmarks and apply Procrustes analysis to gain affine matrix due to the canonical position. Then we perform a similarity transformation to each image and center crop it using a fixed square to get the lip region. By applying such operations on LRW, we find that face alignment is helpful for lip reading, as shown in Tab.\ref{table:dataprocess}. The aligned face video retains yaw and pitch, while removes roll rotation. We infer such operation reducing temporal jittering in face video and encourage the deep model focus on the lip movement rather than pose variations.
\vspace{0.5em}
\\
\indent\textbf{Word Boundary} In 2018, Stafylakis et al. \cite{stafylakis2018pushing} introduced the word boundary by converting it into a binary indicator at each time step and then concatenating it with the original features of the frontend network. We evaluate this manner by the baseline model for lip reading. It shows a significant improvement when introducing this information, shown in Tab.\ref{table:dataprocess}. As described in \cite{stafylakis2018pushing}, RNN has a powerful gating mechanism, and passing the word boundaries variable permits the RNN to make use of them. The out-of-boundaries frames can provide contextual and environmental information (such as the speaker, pose, light, and so on) that might be useful to classify the target word. 

\vspace{-0.5em}
\subsection{Training Tweaks} 
This section introduces several training tweaks for lip reading, which are performed on the baseline model separately to analyze their effects.
\vspace{0.5em}
\\
\indent\textbf{MixUp} To reduce over-fitting, we introduce mixup \cite{zhang2017mixup} as an additional data augmentation method. In this process, two samples A: $(x_A, y_A)$ and B: $(x_B, y_B)$ are selected to generate a new sample $(\hat{x}, \hat{y})$ by a weighted linear interpolation as: 
\begin{align}
	 \hat{x} = \lambda x_A + (1 - \lambda) x_B \notag, \hat{y} = \lambda y_A + (1 - \lambda) y_B \notag
\end{align}
where $x_i, y_i$ denotes the training sample and the word label of data $i\in\{A,B\}$ respectively, $\lambda$ is a number randomly sampled from distribution $Beta(\alpha, \alpha)$. In our implementation, we shuffle each batch $S$ to obtain a second `batch' $S'$ and obtain sample A and B from $S$ and $S'$ respectively. We set $\alpha=0.2$ in our experiments. This operation encourages the deep model to behave linearly in-between training examples, and linearity is a good inductive bias from the perspective of Occam’s razor. As shown in Tab.\ref{table:training}, this augmentation could bring obvious improvements on both datasets. 
\vspace{0.5em}
\\
\indent\textbf{Label Smoothing} 
Given an input sample belonging to word class $i$, we denote $p_i$ as the prediction logits and $y$ as the annotated word label. Let $N$ be the number of classes. Then the traditional cross-entropy loss is computed as:
\begin{equation}
    L= -\sum_{i=1}^{N}q_{i}log(p_{i}) \left\{
\begin{aligned}
q_i=0, y \not=i \\
q_i=1, y=i
\end{aligned}\right.
\end{equation}
When using label smoothing, the construction of $q_i$ is changed as:
\begin{equation}
q_i= \begin{cases}
\epsilon / N & ,y\not=i \\
1-\frac{N-1}{N}\epsilon& ,y=i
\end{cases}
\end{equation}
where $\epsilon$ is a small constant and $\epsilon=0.1$ in our implementation. This operation encourages the model to output a finite probability and generalize better on unseen videos. As shown in Tab.\ref{table:training}, it is advantageous to introduce label smoothing for the word-level lip reading task.
\vspace{-0.5em}
\subsection{Learning Rate Scheduling} 
The learning rate always has a direct impact on the performance of the lip-reading model. 
In our basic pipeline, we reduce the learning rate by a factor of 2 when the validation error meets plateau in 3 continuous epochs. This setting may cause an abrupt reduction of the learning rate. So we also compare with cosine learning rate scheduling and exponential learning rate scheduling. (1) In exponential scheduling setting, the learning rate is multiplied by 0.95 at the end of each epoch. (2) In the cosine setting, the learning rate $\eta_t$ at epoch $t$ is calculated as:
\begin{equation}
\eta_t=\frac{1}{2}(1+cos(\frac{t\pi}{T}))\eta
\end{equation}
where $\eta$ is the initial learning rate. $T$ is the total number of epochs which is 80 in our experiments. The results in Tab. \ref{table:training} show that the cosine scheduling could lead to a small improvement of the performance. In the cosine learning rate schedule, the learning rate decreases slowly at the beginning, almost linear in the middle, and slow again in the end. Compared to other schedulers, the cosine learning rate schedule starts to reduce the learning rate since the start of training while remaining relatively large, which potentially improves the training progress.
\vspace{-0.5em}
\subsection{The Final Pipeline}
Based on the above results, we combine SE and MixUp with the basic pipeline to form our new refined pipeline as in Fig.\ref{fig:architecture} (b). Then, we introduce further the cosine learning rate scheduling, label smoothing, and word boundary  for the learning process. As shown in Tab. \ref{tab:final}, we achieve accuracy of 88.4\% and 55.7\% on LRW and LRW-1000 respectively without changing much of the model, by using only the above refinements. As shown in the table, our results are comparable or even surpass the existing state-of-the-art.
\vspace{-0.5em}
\section{Conclusion}
\vspace{-0.5em}
In this paper, we have performed a comprehensive study on the effects of several factors for lip reading. Without changing much of the main model, we achieved comparable or even better results than the current state-of-the-art. We hope that the comparison and analysis in this study could provide some valuable reference to the related researchers. 
\vfill\pagebreak

\label{sec:refs}

\small{
\bibliographystyle{IEEEbib}
\bibliography{main}}

\end{document}